\documentclass[conference]{IEEEtran}
\IEEEoverridecommandlockouts
\usepackage{balance}
\usepackage{cite}
\usepackage{amsmath,amssymb,amsfonts}
\usepackage{algorithmic}
\usepackage{booktabs}
\usepackage{multirow}
\usepackage{graphicx}
\usepackage{algorithm}
\usepackage{algorithmic}
\usepackage{siunitx}
\usepackage{textcomp}
\usepackage{xcolor}
\def\BibTeX{{\rm B\kern-.05em{\sc i\kern-.025em b}\kern-.08em
    T\kern-.1667em\lower.7ex\hbox{E}\kern-.125emX}}
\newcommand{\AlphaSE}{$\texttt{A}_{\texttt{SE}}$}
\begin{document}

\title{
Smaller Batches, Bigger Gains? Investigating the Impact of Batch Sizes on Reinforcement Learning Based Real-World Production Scheduling
\thanks{This work was partly supported by the Ministry of Economic Affairs, Industry, Climate Action and Energy of the State of North Rhine-Westphalia, Germany, under the project SUPPORT (005-2111-0026)}
}

\author{\IEEEauthorblockN{Arthur Müller} 
\IEEEauthorblockA{\textit{Department of Machine Intelligence} \\
\textit{Fraunhofer IOSB-INA}\\
32657 Lemgo, Germany \\
arthur.mueller@iosb-ina.fraunhofer.de}
\and
\IEEEauthorblockN{Felix Grumbach} 
\IEEEauthorblockA{\textit{Center for Applied Data Science} \\
\textit{Hochschule Bielefeld}\\
33330 Gütersloh, Germany \\
 felix.grumbach@hsbi.de}
\and
\IEEEauthorblockN{Matthia Sabatelli} 
\IEEEauthorblockA{\textit{Department of Artificial Intelligence} \\
\textit{University of Groningen}\\
9712 CP Groningen, The Netherlands \\
 m.sabatelli@rug.nl}
}

\maketitle

\begin{abstract}
Production scheduling is an essential task in manufacturing, with Reinforcement Learning (RL) emerging as a key solution. In a previous work, RL was utilized to solve an extended permutation flow shop scheduling problem (PFSSP) for a real-world production line with two stages, linked by a central buffer. The RL agent was trained to sequence equally-sized product batches to minimize setup efforts and idle times. 
However, the substantial impact caused by varying the size of these product batches has not yet been explored.
In this follow-up study, we investigate the effects of varying batch sizes, exploring both the quality of solutions and the training dynamics of the RL agent. The results demonstrate that it is possible to methodically identify reasonable boundaries for the batch size. These boundaries are determined on one side by the increasing sample complexity associated with smaller batch sizes, and on the other side by the decreasing flexibility of the agent when dealing with larger batch sizes. This provides the practitioner the ability to make an informed decision regarding the selection of an appropriate batch size. Moreover, we introduce and investigate two new curriculum learning strategies to enable the training with small batch sizes. The findings of this work offer the potential for application in several industrial use cases with comparable scheduling problems.

\end{abstract}


\section{Introduction}
\label{sec:introduction}
Reinforcement Learning (RL), a dynamic learning paradigm from the machine learning domain, has gained significant attention over the past decade
The applicability of RL has been explored in various domains including video games~\cite{Mnih2015}, traffic signal control~\cite{Mueller2023}\cite{Noaeen2022}, and robotics~\cite{Brunke2022}. A particularly active area of RL research has been in the field of production scheduling~\cite{Wang2022}\cite{Panzer2022}. The challenges associated with production scheduling manifest in numerous forms across the manufacturing industry, and effectively addressing these challenges can significantly impact the operational costs of manufacturing enterprises~\cite{herrmann2006handbook}.

\begin{figure}
    \centering
    \includegraphics[width=0.48\textwidth]{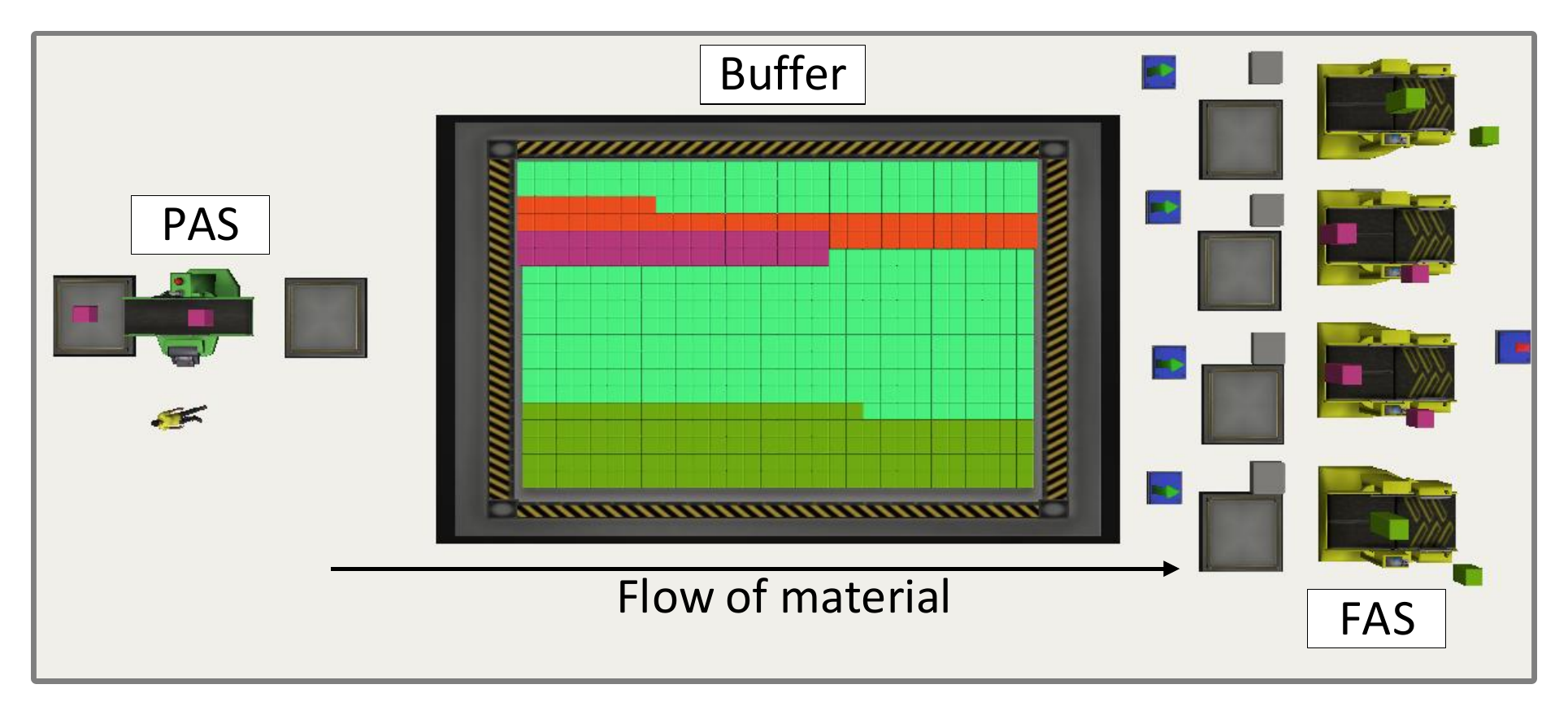}
    \caption{Model of the production system within a discrete-event simulator 
  (adapted from Müller et. al.~\cite{Mueller2024}).}
    \label{fig:sim_model}
\end{figure}

In a previous study~\cite{Mueller2024}, RL was utilized to solve a particular scheduling problem in a real-world production line for the manufacturing of household appliances. 
The production line comprises two stages: a singular preassembly station (PAS) and four final assembly stations (FAS), linked by a central finite buffer. At the PAS, pre-products are produced in batches, stored in the buffer, and then converted into specific products at the FAS. The RL agent learned a policy how to sequence the batches to be manufactured by the PAS in order to minimize idle times and setup efforts, which are conflicting objectives.
Fig.~\ref{fig:sim_model} shows a discrete-event simulation model of this production line, that encapsulates the real-world constraints (see Sect. \ref{sec:pre_problem}) and has been used to train the RL agent (see Sect. \ref{sec:pre_agent}).

The existing RL approach was benchmarked against a genetic algorithm and outperformed it. However, it is important to note that it schedules batches of a fixed size, 
determined from initial practical experiences.
Such a fixed batch size, which has no rigorous empirical basis, may not result in the best outcomes.
Building upon this, in our work we analyze the impact of varying batch sizes on the agent's performance.
Our approach contributes to identifying the optimal batch size ("sweet spot") that balances the agent's training and scheduling efficiency. To the best of our knowledge, the existing literature has not yet addressed this practically relevant issue. 
Moreover, we apply curriculum learning and introduce two new curricula tailored to smaller batch sizes, which are likely to be transferable to comparable real-world production systems. 
An interesting key finding from our experiments indicates that smaller batch sizes result in lower setup efforts but incur longer training times. However, if the batch sizes are too small, it is only possible to train an agent with good scheduling efficiency with the help of the new curricula. But this comes at the price of long training times.

\section{Preliminaries}
\label{sec:preliminaries}

\subsection{Problem formulation}
\label{sec:pre_problem}
The scheduling problem at hand is a modified PFSSP with two stages \cite{Johnson1954}. To succinctly present the problem, it is semi-formally defined as follows:
\begin{itemize}
 \item Consider a set of specific products to be finalized at the FAS, each based on a standardized pre-product of eight different types $\tau$ manufactured at the PAS.
 \item The PAS and all FAS can process only one product at a time and all processing times are deterministic.
 \item The PAS must produce all pre-products demanded by the FAS in the second stage that are not initially available in the central buffer and each pre-product must be finalized at exactly one of four FAS. In this context, each FAS has a predefined schedule containing the sequence of single products to be manufactured.
 \item The PAS produces batches of pre-products of the same type in a uniform batch size $b$. After finishing a whole batch at the PAS, it is put into the buffer, which is limited by a maximum number of pre-products.
 \item Switching from one pre-product type to another may incur sequence-dependent setup efforts $q_{\tau,\tau'}$ at the PAS, where $q_{\tau,\tau'}$ represents the setup effort from product type $\tau$ to $\tau'$.
 \item The FAS can only start finishing the pre-product when it is available in the central buffer. If the next required pre-product is not available in the central buffer when needed, the FAS will incur idle times $d$. It remains idle until the missing pre-product from the first operation becomes available in the buffer.
\end{itemize}

The problem is based on two key decision variables: (1) the uniform batch size $b$ of all batches to be produced at the PAS, and (2) the sequence in which these batches are produced at the PAS. The objective comprises two conflicting metrics to be minimized in a lexicographical order: (1) minimize the idle times of all FAS, and (2) minimize the sequence-dependent setup efforts at the PAS. For ease of reading, pre-products are referred to as products in the following.
\subsection{Reinforcement Learning Approach}
\label{sec:pre_agent}

In this section, we describe the RL solution from the previous study~\cite{Mueller2024}. The scheduling problem is defined as Markov Decision Process (MDP) following standard notation~\cite{Sutton2018}. An MDP is a tuple comprising the following elements:
\begin{itemize}
    \item A set of states $\mathcal{S}$, where a \emph{state} $s_t$ encodes information about the current situation of the MDP at a given discrete time step $t$.
    \item A set of actions $\mathcal{A}$. At each time step $t$, the agent selects an \emph{action} $a_t$ to execute in the environment.
    \item A state transition probability function $\mathcal{P}: \mathcal{S} \times \mathcal{A} \times \mathcal{S} \rightarrow [0, 1]$ that defines a probability distribution for an agent over the successor state $s_{t+1}$ given that an agent is in $s_t$ and selects $a_t$. 
    \item A reward function $\mathcal{R}: \mathcal{S} \times \mathcal{A} \rightarrow \mathbb{R}$ which determines the scalar \emph{reward} $r_{t}$, when an agent selects $a_t$ in state $s_t$.  
    \item A discount factor $\gamma \in [0, 1]$, discounting future rewards.
\end{itemize}
At each time step $t$, the agent selects an action $a_t$ that will be carried out in the environment the agent is interacting with. After taking an action, the environment sends information about its current situation $s_t$ and a scalar reward $r_t$ that represents the immediate reward (or cost) of the action taken. 

\paragraph{State Space}
In the problem, the state is encoded as a vector. For each product type, the vector contains the quantity required for the next \SI{24}{h} as well as the quantity required until the end of the planning horizon, with a plan covering one working week.
In addition, the state contains the remaining time for each product type until its next requirement, denoted as the \textit{range}.
At the beginning of the planning horizon, the buffer is partially filled with products. The quantity of a product type already available in the buffer is taken into account when determining the range and the quantities required. Furthermore, the state encompasses the fill level of the buffer and the last product type produced so that setup effort relationships can be learned.

\paragraph{Action Space}
The action space $\mathcal{A}$ comprises 8 discrete actions ($|\mathcal{A}|=8$) that represent the 8 product types present in this production problem. When an action is selected, the corresponding product type is produced with the predefined batch size $b$. Since the processing time is identical for all products, a constant $b$ over the simulation duration ensures equidistant time steps.
Furthermore, action masking is used to exclude product types from production that are no longer required until the end of the planning horizon. An action mask $m_\text{a}$ contains \texttt{True} for each required type and \texttt{False} otherwise. Actions with a corresponding \texttt{False} cannot be selected by the agent. Empirically, it was shown that action masking reduces the duration of training and increases the probability of successful training.

\paragraph{Reward Function}
A natural approach for defining the reward function would involve using the negatively weighted values of idle time and setup effort.
However, in the previous work, it was demonstrated that reward shaping with respect to idle times simplifies the learning problem significantly. 
Instead of penalizing idle times, the criticality $r_{crit}$ of an product type was used. Criticality is defined as the ratio of the required quantity of an product type within the next 24 hours divided by its range. The larger the required quantity or the smaller the range, the greater the criticality. Additionally, the agent incurs a penalty if the range of a type falls below a threshold of $30$ minutes ($r_\text{mgn}$). We define $r_\text{crit}$ and $r_\text{mgn}$ to be negative values.
This reward shaping approach simplifies the learning problem by avoiding the issue of delayed rewards that would arise from utilizing idle times in the reward function. Poor decisions by the agent would result in idle times only after a certain delay, making it challenging for the agent to identify such bad decisions. Criticality and safety times serve as surrogates for idle times to be expected in the future. By optimizing these surrogate measures, the agent prevents the occurrence of idle times in the first place.
The total reward is given by:
\begin{equation}
\label{eq:reward}
r = r_\text{crit} + r_\text{mgn} - \alpha_\text{se}q_{\tau,\tau'},
\end{equation}
where $\alpha_\text{se}$ represents the penalty factor for setup efforts.

\paragraph{Curriculum Learning}
The learning problem was further simplified by applying curriculum learning. This is a strategy where the agent gradually progresses through a series of tasks of increasing complexity.
In the first two tasks, the agent only learns to optimize for criticality and safety time, the setup efforts are not taken into account ($\alpha_\text{se}=0$). In the first task, a more restrictive action mask $m_\text{easy}$ is used, which only contains the three most urgent product types. In the second task, the "normal" aciton mask $m_\text{a}$ is used, which excludes product types that are no longer required from production. In the last task, the setup effort is taken into account by setting $\alpha_\text{se}$ to \AlphaSE, a scalar value determined in the previous work. A task is considered learned if the sum of idle times per episode $\sum_t d(t)$ is less than $100s$ on average over the last 100 episodes. This is the necessary condition for the transition to the next task in the curriculum: 
\begin{equation}
\label{eq:transition_condition}
    \frac{1}{100} \sum_{i=1}^{100} \left( \sum_t d_i(t) \right) < 100s
\end{equation} 
The curriculum is summarized in Table~\ref{tab:curriculum1}, referred to as Curriculum~A.

\begin{table}[htbp]
\centering
\caption{Curriculum A - Base Curriculum}
\begin{tabular}{c|c|c}
Task & Action Mask & \( \alpha_\text{se} \) \\
\midrule
1 & \( m_{\text{easy}} \)   & $0$ \\
2 & \( m_{\text{a}} \) & $0$ \\
3 & \( m_{\text{a}} \) & \AlphaSE \\
\end{tabular}
\label{tab:curriculum1}
\end{table}

\section{Decreasing the Batch Size}
In the previous study, all experiments were conducted with a fixed batch size of $50$ products, a value determined from initial trials rather than through comprehensive analysis.
But, if, for example, 51 products of a type still are required, 49 products would have to be built and put in the buffer that are not immediately required. This would result in other, more urgent types not being built during this time. In addition, the buffer could be filled with types that may not be needed until very late.
Smaller batch sizes, therefore, are more desirable with regard to the optimization problem, as the objectives can be targeted more precisely.

\subsection{Experimental Setup}
This motivates us to train the agent with smaller batch sizes. We conduct an experiment with batch sizes of $10$, $20$, $30$, and $40$ products.
For comparison and deeper understanding, we also increase the batch size and train agents with batch sizes from $50$ up to $100$. 
Each setting is trained 10 times according to Curriculum~A. We use the Proximal Policy Optimization (PPO) algorithm~\cite{Schulman2017} as RL algorithm because it has proven to be more robust and less hyperparameter sensitive than other algorithms for this problem~\cite{Mueller2024}. 
The maximum number of time steps is set to \(5 \cdot 10^6\).
The final criterion for the successful completion of training, once the agent is in the last task, is set to be stricter than the condition specified in Equation~\ref{eq:transition_condition}:
\begin{equation}
    \frac{1}{100} \sum_{i=1}^{100} \left( \sum_t d_i(t) \right) < 15s
\end{equation} 
We use a dataset from the previous work known for its high combinatorial complexity, making it challenging for the agent to solve.

Reducing the batch size increases the number of time steps per episode. However, this would also affect the sum of rewards per episode (\textit{return}), so that, for example, the return for an agent with a batch size of 50 and one with a batch size of 10 would be different, given identical behavior. This is because an agent with a batch size of 10 would collect an approximately $50/10=5$ times higher value for the sum of the reward components $r_\text{crit}$ and $r_\text{mgn}$ in the same period. To maintain almost identical returns for identical agent behavior, we therefore normalize them for all batch sizes $b$ to that of 50:

\begin{equation}
\label{eq:reward_crit}
r'_\text{crit} = r_\text{crit} \frac{b}{50}
\end{equation}

\begin{equation}
\label{eq:reward_mgn}
r'_\text{mgn} = r_\text{mgn} \frac{b}{50}
\end{equation}

Thus, we get the new adjusted reward $r'$: 

\begin{equation}
\label{eq:reward_function}
r' = r'_\text{crit} + r'_\text{mgn} -\alpha_\text{se}q_{\tau,\tau'}
\end{equation}

For evaluation, we consider both the training metrics and the solution quality. Since the inference of a trained agent is cheap, we let each trained agent generate a plan 20 times with a stochastic policy and select the best plan that has the lowest setup efforts at the smallest idle time. In our experience, the solutions based on a stochastic policy are often slightly better than those based on a deterministic policy, which is consistent with the empirical results reported in~\cite{Wu2022}.

\subsection{Results and Discussion}

Table~\ref{tab:results} provides an overview of the results. It is notable that the agents with the smallest and largest batch sizes performed poorly.
Specifically, for a batch size of 10, no agent met the final criterion of Curriculum~A, nor were they able to generate zero-idle-time plans. At a batch size of 100, only in one instance was an agent able to generate a plan with no idle times. We will discuss the reasons for these results later on in this section. For all other batch sizes, the training runs were successful, meaning that the agents fulfilled the lexicographic optimization goal by first minimizing idle times and then setup efforts.

\begin{table*}[h!]
  \centering
  \caption{Results for Curriculum A with different batch sizes. Each setting is trained 10 times.}
  \label{tab:results}
  \begin{tabular}
  {
    S[table-format=2.0]
    |S[table-format=2.0]
    |S[table-format=2.0]
    |S[table-format=1.2e+2]
    S[table-format=1.2e+2]
    S[table-format=1.2e+2]
    |S[table-format=3.0]
    S[table-format=3.0]
    S[table-format=3.0]
    S[table-format=2.1]
    |S[table-format=-3.0]
    S[table-format=-3.0]
    S[table-format=-3.0]
  }
    \multicolumn{1}{c|}{\multirow{2}{*}{\shortstack{batch\\size}}} & \multicolumn{1}{c|}{\multirow{2}{*}{\shortstack{final\\ criterion met}}} & \multicolumn{1}{c|}{\multirow{2}{*}{\shortstack{zero-idle-\\time solution}}} & \multicolumn{3}{c|}{time steps needed for finishing training\textsuperscript{1}} & \multicolumn{4}{c|}{setup efforts\textsuperscript{2}} & \multicolumn{3}{c}{return\textsuperscript{2}} \\
     &  &  & {min} & {avg} & {max} & {min} & {avg} & {max} & {std} & {min} & {avg} & {max} \\
    \midrule
    {$10$} & {$0$} & {$0$} & {--} & {--} & {--} & {--} & {--} & {--} & {--} & {--} & {--} & {--} \\
    {$20$} & {$8$} & {$10$} & {$1.16\cdot 10^6$} & {$3.44\cdot 10^6$} & {$4.88\cdot 10^6$} & {$147$} & {$191$} & {$239$} & {$28.8$} & {$-429$} & {$-362$} & {$-293$} \\
    {$30$} & {$10$} & {$10$} & {$7.33\cdot 10^5$} & {$1.70\cdot 10^6$} & {$3.38\cdot 10^6$} & {$182$} & {$211$} & {$237$} & {$14.4$} & {$-367$} & {$-330$} & {$-312$} \\
    {$40$} & {$10$} & {$10$} & {$7.20\cdot 10^5$} & {$1.23\cdot 10^6$} & {$2.20\cdot 10^6$} & {$184$} & {$208$} & {$246$} & {$17$} & {$-371$} & {$-348$} & {$-321$} \\
    {$50$} & {$10$} & {$10$} & {$5.82\cdot 10^5$} & {$7.98\cdot 10^5$} & {$1.23\cdot 10^6$} & {$219$} & {$228$} & {$238$} & {$6.0$} & {$-339$} & {$-319$} & {$-305$} \\
    {$60$} & {$10$} & {$10$} & {$5.13\cdot 10^5$} & {$7.81\cdot 10^5$} & {$1.16\cdot 10^6$} & {$228$} & {$239$} & {$249$} & {$6.1$} & {$-349$} & {$-338$} & {$-329$} \\
    {$70$} & {$10$} & {$10$} & {$4.92\cdot 10^5$} & {$6.32\cdot 10^5$} & {$7.66\cdot 10^5$} & {$219$} & {$227$} & {$235$} & {$5$} & {$-347$} & {$-328$} & {$-317$} \\
    {$80$} & {$10$} & {$10$} & {$4.51\cdot 10^5$} & {$6.46\cdot 10^5$} & {$8.40\cdot 10^5$} & {$251$} & {$258$} & {$267$} & {$4.8$} & {$-410$} & {$-393$} & {$-384$} \\
    {$90$} & {$10$} & {$10$} & {$4.68\cdot 10^5$} & {$7.20\cdot 10^5$} & {$9.93\cdot 10^5$} & {$250$} & {$263$} & {$282$} & {$9.2$} & {$-412$} & {$-403$} & {$-396$} \\
    {$100$} & {$0$} & {$1$} & {--} & {--} & {--} &  & {$275$} &  & {--} &  & {$-435$} &  \\
    \bottomrule
  \end{tabular}
   \par
  \vspace{5pt} 
  \small 
  \textsuperscript{1}Considering only solutions that met the final criterion of the curriculum.\par
  \textsuperscript{2}Considering only zero-idle-time solutions.
\end{table*}

\begin{figure}
    \centering
    \includegraphics[width=0.48\textwidth]{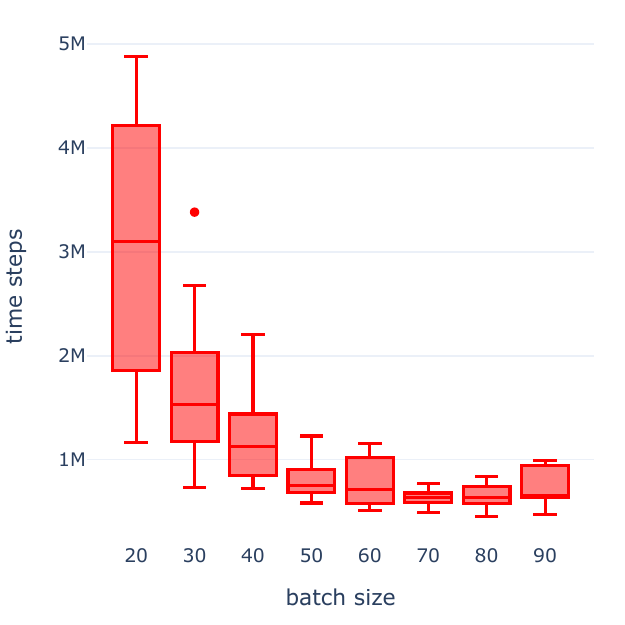}
    \caption{Time steps needed to finish the training. Two separate effects cause an increase: As the batch size decreases, the sample complexity increases due to the increasing policy space. As the batch size increases, the sample complexity increases due to the decreasing flexibility of the agent, which makes it harder to find good solutions. The minimum is at $b=70$.}
    \label{fig:boxplot_trainingtime}
\end{figure}

\begin{figure}
    \centering
    \includegraphics[width=0.48\textwidth]{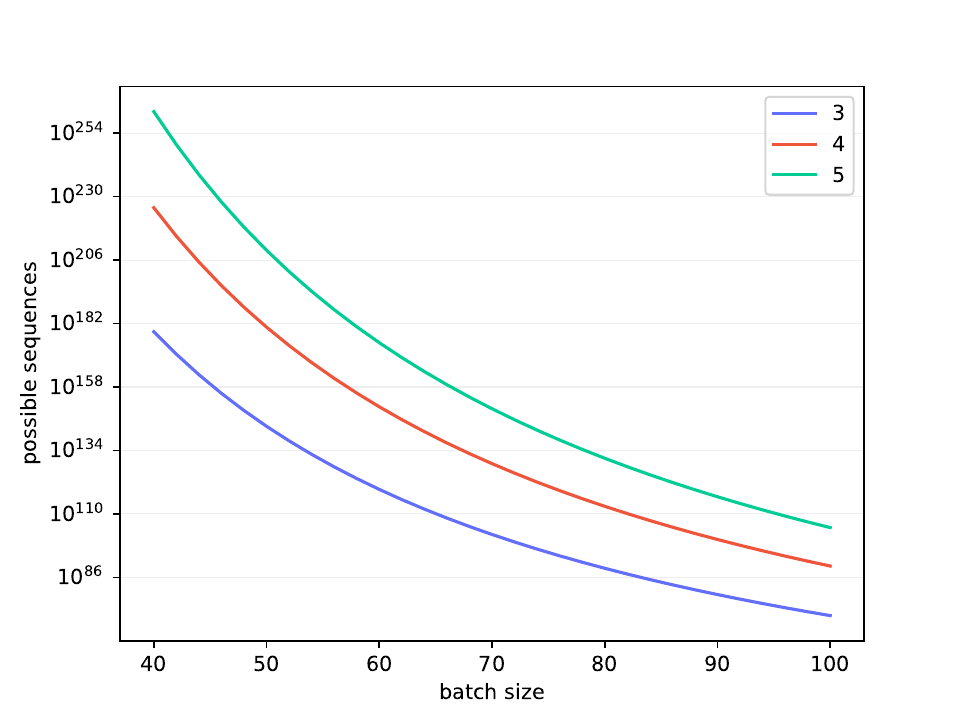}
    \caption{$|\mathcal{A}|^T$ over batch sizes for different average number of actions per step. $|\mathcal{A}|=8$, $T=300$ for $b=50$ and a planning horizon of one week. The y-axis is presented on a logarithmic scale. Due to limited computational capability, we could not perform calculations for values smaller than $b=40$.}
    \label{fig:AhochT}
\end{figure}

Next, we examine the required number of time steps to complete the training for all agents that met the final criterion (see Fig.~\ref{fig:boxplot_trainingtime}).
According to~\cite{Laidlaw2024}, the median of the number of time steps to solve a problem is an estimate of the sample complexity of the RL algorithm.
At a batch size of 70, the median sample complexity is at its lowest point $6.4\cdot10^5$. As the batch size decreases, the sample complexity increases, reaching a median of $3.1\cdot10^6$ at $b=20$.
This is due to the exponential growth of the possible number of sequences (\textit{policy space}) that can be calculated by $|\mathcal{A}|^T$, where $T$ represents the number of time steps per episode. With decreasing batch size, the number of time steps per episode increases ($T \propto \frac{1}{b}$) for the same planning horizon.
We illustrate this dramatically expanding policy space in Fig.~\ref{fig:AhochT}, for an average number of actions per step ranging from 3 to 5 (determined empirically).
This expanded policy space, resulting from smaller batch sizes, makes the problem more difficult to solve with RL, which is why the algorithm requires more samples to solve it (see also~\cite{Furuta2021}).

However, starting from a batch size of 70, the sample complexity increases slightly with an increasing batch size, even though the policy space actually shrinks.
Once the batch size exceeds a certain threshold, idle times become unavoidable as the agent loses the necessary flexibility to respond dynamically enough to the demands of the FAS. 
For that reason, with a batch size of 100, the potential solution space is apparently so limited that a zero-idle-time solution is only achieved in one out of ten runs. 
From this, we can derive a general principle regarding the optimal number of time steps needed to complete training: two opposing effects must be taken into account. An increasing policy space with decreasing batch size and a decreasing agent flexibility with increasing batch size.

\begin{figure}
    \centering
    \includegraphics[width=0.48\textwidth]{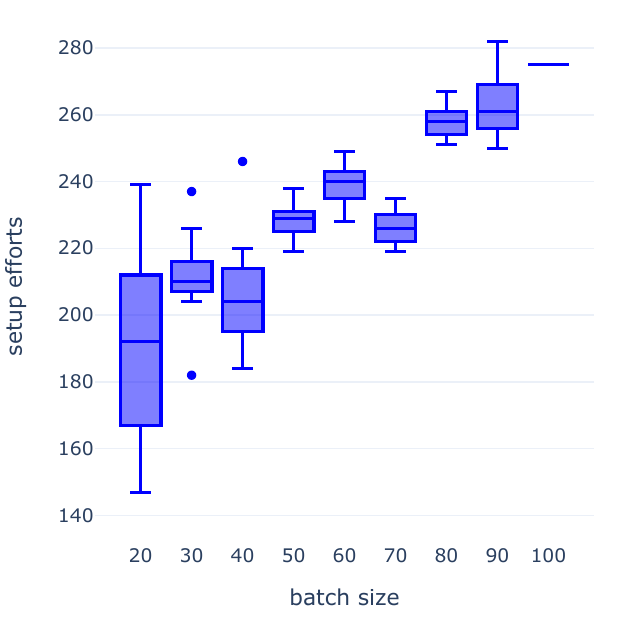}
    \caption{Setup efforts for batch sizes. Considering only zero-idle-time solutions.}
    \label{fig:boxplot_setupefforts}
\end{figure}

Now, we turn to a deeper analysis of the setup efforts, see Fig.~\ref{fig:boxplot_setupefforts}. In general, the median setup effort decreases as the batch size decreases, ranging from $275$ at a batch size of $100$ to $192$ at a batch size of $20$. 
However, there are exceptions to this trend. For instance, the median for a batch size of 40 is lower than that of 30, and the median for a batch size of 70 is even lower than that of 60 and 50. It is possible that for the specific dataset under consideration, these batch sizes may offer favorable optimization opportunities.
One reason for decreasing setup efforts with decreasing batch sizes may be partly because the greater flexibility through smaller batch size let the agent find policies that better meet the objectives.
Another reason is that the penalties for setup efforts have a greater impact when batch sizes are smaller. When the agent enters task 3 of Curriculum~A - the agent should now learn to minimize setup efforts ($\alpha_{se}$=\AlphaSE) - it initially performs more type changes with smaller batch sizes than with larger ones. 
The relative contribution of the setup effort penalty to the reward is therefore higher for smaller batch sizes.
For this reason, at the beginning of task 3, there is a greater focus on avoiding setup effort for smaller batch sizes. Accordingly, it is more likely that policies are learned that accept a higher criticality ($r_{crit}$) and a more frequent drop below the 30 min threshold ($r_{mgn}$) in favor of setup effort minimization ($\alpha_\text{se}q_{\tau,\tau'}$).
This can also be seen when comparing returns and setup efforts in Tab.~\ref{tab:results}. On average, better returns are achieved with a batch size of $50$ to $70$ than with one in the range of $20$ to $40$. But these better returns comes with higher setup efforts. This can only be explained by the fact that a higher criticality and a more frequent drop is permitted when the batch sizes decrease.

Furthermore, it can be observed that in general the standard deviation of the setup efforts increases with decreasing batch size, rising from around 5-6 at a batch size of $50-80$ to 28.8 at $b=20$. As the policy space grows exponentially with the reduction in batch size, it becomes increasingly likely for the agent to discover diverse policies, some of which may be more or less efficient than others. 
For this reason, the worst solution with a batch size of 20 (239) is even slightly worse than the worst solution with a batch size of 50 (238).

From the observations made, the key findings for practitioners in determining the optimum batch size can be summarized as follows:
\begin{enumerate}
    \item Smaller batch sizes increase the probability that the agent will find plans with lower setup efforts. However, this also leads to an exponential increase in sample complexity and, consequently, the required training duration. It should also be noted that as the batch size decreases, the variance in setup effort increases significantly, potentially leading to mediocre solutions. In these cases, accepting a longer training period does not lead to any advantages.
    \item Batch sizes that are too small cause the training to fail and should therefore be avoided.
    \item There exists a minimum for sample complexity. In our case, this minimum occurs at a batch size of 70. Batch sizes larger than this point should be avoided, as they not only increase sample complexity but also cause higher setup efforts, offering no advantage.    
    \item Therefore, a reasonable range for batch size is bounded on one end by the point beyond which training becomes feasible and on the other by the minimum of sample complexity. 
    Within this range, practitioners must find a balance between setup effort and sample complexity. This balance depends on the computing resources available for potentially parallelizing the training and making it faster, the time available in the production for solving the optimization problem, and the actual costs incurred from increased setup.
    \item There are batch sizes that have a particularly low setup effort in proportion to the required sample complexity (in our case $b=40$ and $b=70$). We suspect that this is related to the specific data set used. Further research is needed in this area to possibly identify these batch sizes a-priori.
\end{enumerate}

\subsection{Analyzing the Challenges of Small Batch Sizes}
\begin{figure}
    \centering
    \includegraphics[width=0.48\textwidth]{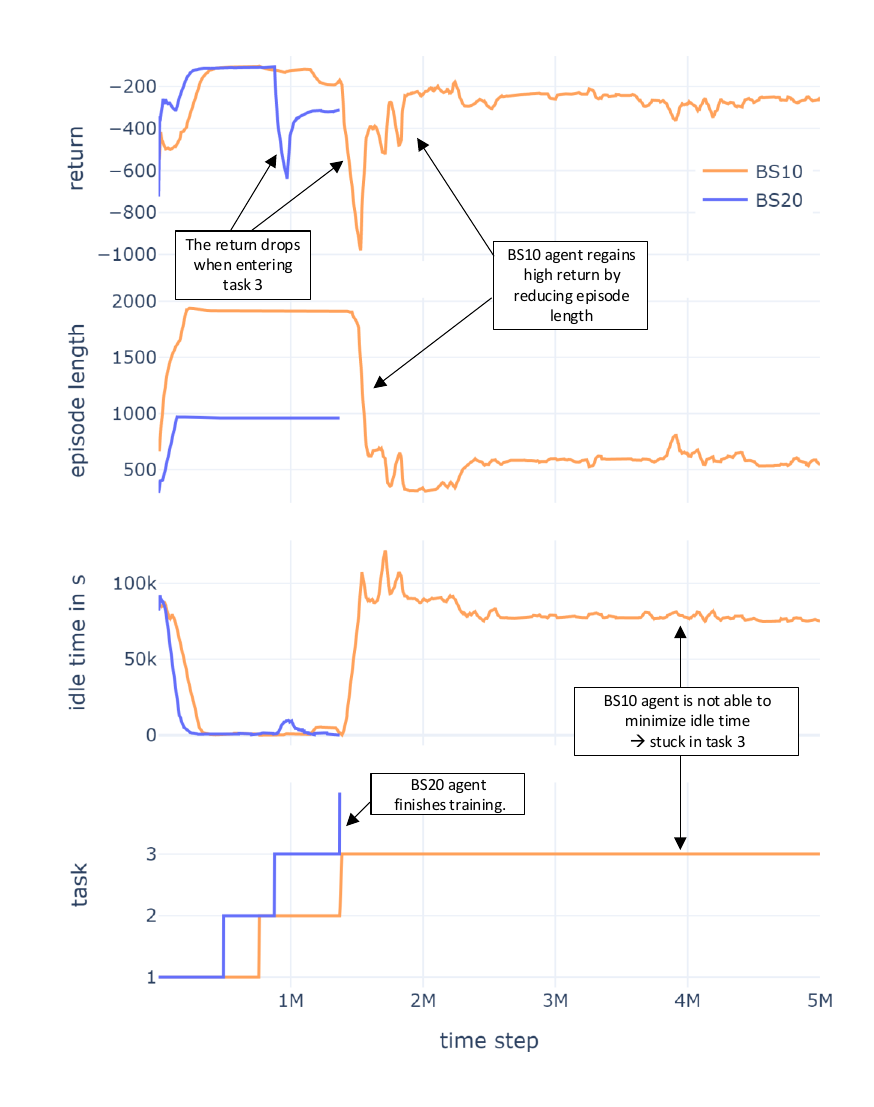}
    \caption{Learning curve for a batch size of 10 (BS10 agent) and 20 (BS20 agent). The BS10 agent is drifting away from the intended behavior when entering task 3.}
    \label{fig:BS10problem}
\end{figure}

To investigate why training with a batch size of 10 is failing, we now perform a deeper analysis. Fig.~\ref{fig:BS10problem} shows various metrics of the learning curve for a batch size of 10 (BS10 agent) and 20 (BS20 agent). Both agents are capable of learning the first two tasks (reducing idle time) and achieve a roughly equal level of return. Upon entering task 3, a drop in return is observed for both agents due to the added penalty for setup effort. This drop is approximately twice as significant for the BS10 agent since this agent makes twice as many decisions that, in an untrained state, result in approximately twice the setup effort. However, after a certain period, the agent in both cases learns to again reach a higher return level, at which it stabilizes. But the BS10 agent irreversibly drifts away from its initially good policy, which reduced idle times. The agent learns instead to reduce the episode length by filling the buffer with unused products, causing the system to enter a deadlock and end prematurely. This results in an increase in the return level by collecting fewer negative rewards overall, due to fewer time steps per episode. The actual desired behavior can be seen in the BS20 agent. After a drop, the return recovers without increasing the idle times or shortening the episode length. In other words, the return is recovered by minimizing the setup effort, which is the intended behavior. We use these findings to develop new curricula that are better able to deal with small batch sizes.

\section{Improving the Learning for Small Batch Sizes}
For our RL approach, a batch size that is too small makes the problem unsolvable. Merely extending the training duration does not resolve this issue, as the policy deviates irreversible from the desired behavior when transitioning from task 2 to task 3, because it learns to reduce the episode length. To facilitate training with a batch size smaller than 20 — potentially leading to reduced setup efforts — we have developed two alternative curricula that simplify the learning problem. Both curricula are designed to smooth the transition from task 2 to task 3, thereby preventing the policy from prematurely entering a deadlock. We trained agents using both curricula ten times each, allowing for a maximum number of $15\cdot10^6$ time steps, as it was evident from the previous experiment that $5\cdot10^6$ time steps would be insufficient. Both curricula, Curriculum B and C, are presented in Table~\ref{tab:curriculum2and3}.

\subsection{Curricula for Small Batch Sizes}
Curriculum B is designed to enable training with a batch size as low as 10. It starts with the first two tasks from Curriculum A, which focus solely on reducing idle times without considering setup efforts ($\alpha_{SE}=0$). Then, the penalty for setup efforts is incrementally increased in $20\%$ steps of \AlphaSE~until it is fully integrated at $100\%$\AlphaSE. Contrary to Curriculum A, this approach divides task~3 into five sub-tasks, each placing an increasingly greater emphasis on minimizing setup efforts. This gradual approach aims to prevent the policy drift observed in Fig.~\ref{fig:BS10problem} by avoiding a large drop in the return.

Conversely, Curriculum C begins with a batch size of 20—the smallest batch size that previously resulted in successful training outcomes. After completing all three tasks from Curriculum A, in which the agent has learned to minimize both setup efforts and idle times, the batch size is gradually reduced by 2 for each subsequent task until it reaches 10. The idea behind this approach is that with only small reductions in batch sizes, the agent's policy does not change too drastically, thus maintaining the desired behavior.

\begin{table}[htbp]
\centering
\caption{Curricula to improve the learning for small batch sizes}
\begin{tabular}{cc|cc|cc}
\toprule
 &  & \multicolumn{2}{c|}{Curriculum~B} &  \multicolumn{2}{c}{Curriculum~C} \\
task & action mask & \( b \) & \( \alpha_\text{se} \) & \( b \) & \( \alpha_\text{se} \) \\
\midrule
1 & \( m_{\text{easy}} \)   & 10 & 0     & 20 & 0     \\
2 & \( m_{\text{a}} \) & 10 & 0     & 20 & 0     \\
3 & \( m_{\text{a}} \) & 10 & $0.2\cdot$\AlphaSE & 20 & \AlphaSE     \\
4 & \( m_{\text{a}} \) & 10 & $0.4\cdot$\AlphaSE & 18 & \AlphaSE     \\
5 & \( m_{\text{a}} \) & 10 & $0.6\cdot$\AlphaSE & 16 & \AlphaSE     \\
6 & \( m_{\text{a}} \) & 10 & $0.8\cdot$\AlphaSE & 14 & \AlphaSE     \\
7 & \( m_{\text{a}} \) & 10 & $1.0\cdot$\AlphaSE     & 12 & \AlphaSE     \\
8 & \( m_{\text{a}} \) & -  & -     & 10 & \AlphaSE     \\
\end{tabular}
\label{tab:curriculum2and3}
\end{table}


\subsection{Results and Discussion}
In Table~\ref{tab:cur2vscur3}, the results are presented. For each run, we use the agent at the point when it had just successfully completed the last task. As in the previous experiment, each agent is evaluated 20 times with a stochastic policy, and the plan selected is the one with that has the lowest setup efforts at the smallest idle time.
It is notable that no agent trained under Curriculum B met the final criterion, whereas only one agent following Curriculum C achieved this (task 8 completed).
To understand these results, we examined the training metrics for all agents, similar to the analysis presented in Fig.~\ref{fig:BS10problem}, focusing on whether policies drifted irreversibly after task transitions.
The findings indicate that policies trained under Curriculum B exhibited this drifting behavior nine times, while those trained under Curriculum C experienced it four times.
This means that despite the smoother transitions of these curricula, irreversible policy drifts occur in many cases at a certain point. 

\begin{figure}
    \centering
    \includegraphics[width=0.2\textwidth]{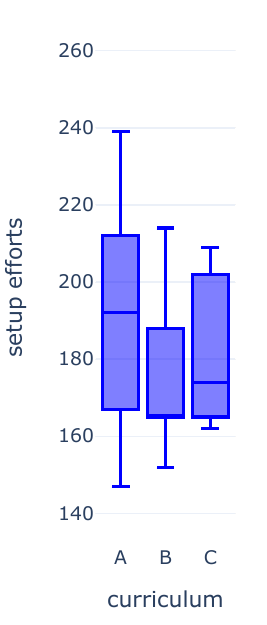}
    \caption{Setup efforts for Curriculum A ($b=20$), B and C.}
    \label{fig:boxplot_CurBC}
\end{figure}

We focus now on comparing the setup efforts achieved under the new curricula with those from Curriculum A with a batch size of 20.
As depicted in Figure~\ref{fig:boxplot_CurBC}, both Curriculum B and C resulted in lower median and variance of setup efforts compared to Curriculum A, with respective values for median setup efforts being 192 for Curriculum A, 165 for B, and 174 for C. 
Curriculum B performed better than C in terms of lower median setup effort, although with a slightly higher standard deviation (19.1 vs. 18.7). Surprisingly, Curriculum B achieved such a low setup effort even without fully reaching the penalty factor at $100\%$\AlphaSE. This supports our previous finding that the relative contribution of setup effort penalty to the reward is higher for smaller batch sizes initially, leading to strategies that seek to minimize setup effort more than for higher batch sizes.
It is also worth noting that the best setup effort was achieved under Curriculum A.

From a practical point of view, if sufficient time and computational resources are available, then the new curricula offer useful alternatives to Curriculum~A. Curriculum~B appears to be slightly better than C, although further experimentation with different datasets is needed to confirm this conclusion. It is important to note, however, that the gains in reducing setup effort are relatively modest when compared to the threefold increase in the maximum allowed training duration. Furthermore, because of the high variance in the resulting setup efforts, there is no guarantee that the new curricula will outperform Curriculum A.

\begin{table}[h!]
  \centering
  \caption{Performance of all agents trained with Curriculum~B and C, sorted by the last task reached}
  \label{tab:cur2vscur3}
  \begin{tabular}{ccc|ccc}
    \toprule
    \multicolumn{3}{c|}{\textbf{Curriculum~B}} & \multicolumn{3}{c}{\textbf{Curriculum~C}} \\
    \multirow{2}{*}{\shortstack{setup\\effort}} & \multirow{2}{*}{\shortstack{idle\\time}} & \multirow{2}{*}{\shortstack{task\\completed}}  & \multirow{2}{*}{\shortstack{setup\\effort}} & \multirow{2}{*}{\shortstack{idle\\time}} & \multirow{2}{*}{\shortstack{task\\completed}} \\ [10pt]
    \midrule
    198 & 0 & 4 & 206 & 0 & 3 \\
    163 & 0 & 4 & 162 & 0 & 3 \\
    214 & 0 & 5 & 202 & 0 & 5 \\
    188 & 0 & 5 & 180 & 0 & 5 \\
    175 & 0 & 5 & 209 & 0 & 6 \\
    165 & 0 & 5 & 172 & 0 & 6 \\
    166 & 0 & 6 & 165 & 0 & 6 \\
    165 & 0 & 6 & 176 & 0 & 7 \\
    165 & 0 & 6 & 165 & 0 & 7 \\
    152 & 0 & 6 & 163 & 0 & 8 \\
  \end{tabular}
\end{table}

\section{Conclusion}
\label{sec:conclusion}

The starting point of this study was an RL-based scheduling system developed for a real-world household appliance production. This system has so far focused on operating with a predefined batch size based on initial trials. To explore the potential benefits of using different batch sizes in such realistically constrained production lines, we rigorously investigated their impact on both the RL training process and the quality of the resulting plans. The results show that smaller batch sizes lead to reduced setup efforts. This is because the setup effort penalty has a greater impact on the reward during the training, which encourages policies that focus more on minimizing setup efforts. Furthermore, smaller batch sizes enable the generation of plans that meet the defined objectives more precisely.
However, if the batch size gets too small, the problem becomes unsolvable for the RL approach. This is due to a sharp transition that occurs with the original curriculum learning strategy. To address this issue, we have introduced two new curricula that smoothen this transition, resulting in better setup efforts on average.

In addition, our research revealed that there is an optimal batch size at which the number of time steps required for training is minimal. As the batch size decreases from this point, the training time increases due to an exponentially larger policy space. If the batch size increases from this point, this limits the flexibility of the agent, which also increases the training time. This point therefore defines the upper limit of a reasonable range, as larger batch sizes lead to both longer training times and higher setup efforts. Our work helps practitioners determine the optimal batch size.

We believe the results of our work may be applicable to similar scheduling problems, but additional research is required to validate this assumption. A promising direction for future studies is to further develop the curricula and devise new learning strategies that make the training process for small batch sizes more stable and efficient.

\balance
\bibliography{references}
\bibliographystyle{IEEEtran}

\end{document}